# Comparison of Two Augmentation Methods in Improving Detection Accuracy of Hemarthrosis

Qianyu (Rita) Fan

qianyu.fan@mail.utoronto.ca

Department of Medical Imaging, University of Toronto

# Abstract

**Introduction:** With the increase of computing power, machine learning models in medical imaging have been introduced to help in rending medical diagnosis and inspection, like hemophilia, a rare disorder in which blood cannot clot normally. Often, one of the bottlenecks of detecting hemophilia is the lack of data available to train the algorithm to increase the accuracy. As a possible solution, this research investigated whether introducing augmented data by data synthesis or traditional augmentation techniques can improve model accuracy, helping to diagnose the diseases.

**Methods:** To tackle this research, features of ultrasound images were extracted by the pre-trained VGG-16, and similarities were compared by cosine similarity measure based on extracted features in different distributions among real images, synthetic images, and augmentation images (Real vs. Real, Syn vs. Syn, Real vs. Different Batches of Syn, Real vs. Augmentation Techniques). Model testing performance was investigated using EffientNet-B4 to recognize "blood" images with two augmentation methods. In addition, a gradient-weighted class activation mapping (Grad-CAM) visualization was used to interpret the unexpected results like loss of accuracy.

**Results:** Synthetic and real images do not show high similarity, with a mean similarity score of 0.4737. Synthetic batch 1 dataset and images by horizontal flip are more similar to the original images. Classic augmentation techniques and data synthesis can improve model accuracy, and data by traditional augmentation techniques have a better performance than synthetic data. In addition, the Grad-CAM heatmap figured out the loss of accuracy is due to a shift in the domain.

**Conclusion:** Overall, this research found that two augmentation methods, data synthesis and traditional augmentation techniques, both can improve accuracy to a certain extent to help to diagnose rare diseases.

# Introduction

*Background*

Hemarthrosis occurs after physical trauma or in bleeding disorders especially hemophilia [1], a rare hemorrhagic disorder caused by inactive or deficient blood-clotting proteins that results in a less efficient coagulation system [2]. Pain follows with swelling of the joints at the beginning, and if it progresses, hemarthrosis makes cartilage thinner and more inflamed, bringing on weakness and additional bleeding into joints. Even worse, permanent damage in joint structure and function would be created [3]. Hemophilia is not only a regional disease, but also a global hazard. A study showed that about 20,000 people have hemophilia in the United States, and 400,000 worldwide live with it [4]. Thereby, diagnosing hemarthrosis is a significant and meaningful topic.

Under-diagnoses may happen in detecting hemophilic arthropathy since doctors rely on previous experience and subjective assessments, causing early changes to be ignored. The lack of prompt identification and appropriate treatment are inclined to a more challenging clinical scene [2, 5]. With the increase of computing power, machine learning in medical imaging has been introduced to help in rending medical diagnoses. However, there are still some limitations. One of the bottlenecks is the lack of data available to train the algorithm, not to mention if one also needs labeled data [6]. Machine learning models in analyzing medical imaging require a vast number of images as training datasets that are both costly and not always available, especially for rare diseases. On the other hand, the model used in training is not good enough for disease identification. To optimize these issues, increasing the scope of training dataset is a common way.

There are two methods to overcome the conundrum of data availability: data synthesis and traditional augmentation techniques. Synthetic data is generated by computer algorithms as a substitute for original data. The benefits of introducing synthetic data are cost-effective and anonymous [6]. The other method is traditional augmentation techniques, which significantly expands the size and diversity of data available for training models. The most popular and effective practice is image transformations, including rotation, reflection, scaling, and shearing [7]. The main difference between data synthesis and traditional augmentation techniques is whether to inject new information in the training dataset. Data synthesis is creating new images and adding them to the training dataset, but traditional augmentation techniques artificially increase the size of the training dataset by transforming the original data.

The purpose of this research was to compare two augmentation methods, data synthesis and traditional augmentation techniques, by measuring the effect on testing model accuracy. We will use both data synthesis and traditional augmentation techniques to increase the size of the training dataset.

*Research Question*

Whether introducing augmented data by using data synthesis or traditional augmentation techniques can improve model accuracy in the training process?

*Hypotheses*

In this paper, it is hypothesized that 1) Data synthesis can improve the accuracy; 2) Traditional augmentation techniques can improve the accuracy; and 3) Synthesis can improve accuracy with better performance than traditional augmentation techniques in the training process.

*Objectives*

1. Describe and compare the similarity between augmented data by data synthesis and traditional augmentation techniques with original images by using the cosine similarity measure
2. Explore the recognition accuracy of image datasets when adding augmented data with either data synthesis or traditional augmentation techniques with increasing proportions of augmented data
3. Compare the recognition accuracy of both augmentation methods over varying proportions of augmented data

## Methods

*Dataset*

The dataset used in this research consists of 658 real hemophilia knee ultrasound images where 336 are images containing recess distension with blood while the remaining 322 are images without blood, and 935 synthetic hemophilia images where 467 are images containing recess distension with blood and 468 are images without blood. Three batches are categorized in synthetic images -- batch 1 has 47 images, batch 2 has 292 images, and batch 3 has 596 images -- each batch has different levels of similarity to real images. The real images are acquired from suspected patients in community clinics, and the synthetic images are generated using SofTx Innovations.

*Feature Extraction*

A pre-trained VGG16 model is used for feature extraction. The VGG-16 consists of 16 weight layers, and the input layer takes each image in the size of (224x224x3). The feature extraction part of the model is from the input layer to the last max pooling layer (7x7x512), and fully connected layers (1x1x4,096) compile the extracted data [8]. Thereby, we used the VGG-16 without loading the last two layers, and input images were transformed into an array with 4, 096 vectors (see Figure 1).

*Similarity Comparison*

Cosine similarity is one of measurements of similarity between two non-zero vectors. Nguyen and Bai [9] used cosine similarity metric for face verification, so we also used cosine similarity metric to measure the similarity between two images.

Cosine similarity (CS) between two vectors $x$ and $y$ is defined as:

$$CS(x,y) = \frac{xy}{||x|| \, ||y||} = \frac{\sum_{i=1}^{n} x_i y_i}{\sqrt{\sum_{i=1}^{n} x_i^2} \sqrt{\sum_{i=1}^{n} y_i^2}},$$

where $x_i$ and $y_i$ are feature vectors of $x$ and $y$. The resulting similarity measure is always within the range of -1 and 1. The score of two identical images is 1 and -1 for two extremely different images.

*Image Recognition and Classification*

EfficientNet, a state-of-art convolutional neural network, is used for image recognition and classification. It works on the idea that providing an effective compound scaling method – scaling all dimensions of depth, width, and resolution – for increasing the model size and attaining better accuracy. [10] illustrates that the EfficientNet models outperform over existing convolutional neural networks, with reduced parameter size. To balance the efficiency and accuracy, we chose EfficientNet-B4 for image recognition and classification.

*Visual Explanation*

Visualization techniques are utilized to comprehend the area of interest that the neural network uses to make the decision. Gradient-weighted class activation mapping (Grad-CAM) is a heatmap technique which uses gradients from trained neural network to produce a coarse localization map highlighting the vital regions of image for predicting the image's classification [11], so this heatmap can help us to interpret some unexpected results made by neural network's prediction.

*Experiment Environment*

Experiments were run using the standard Python 3.8.6 interpreter in MacBook Pro 13-inch with 2.4 GHz Intel Core i5. All image preprocessing and feature extraction were done with TensorFlow 2.3.1, and we used PyTorch 1.9.0 to train the model and made Grad-CAM heatmaps. The results we got were run in GPU Nvidia GeForce 2080 Ti and plotted in R 4.1.1 programming language.

*Experimental Process*

To assess how similar are the augmented data by data synthesis to the real images, we first predicted features of the ultrasound images by the pre-trained VGG-16. The cosine similarity scores were calculated based on their extracted features for comparing real images to synthetic images with blood and without blood, respectively. We then replicated the steps to check how similar are the augmented data by traditional augmentation techniques -- 90 degrees clockwise rotation, horizontal flip, and contrast enhancement -- to the real images (see Appendix).

In order to investigate whether introducing augmented data by using data synthesis or traditional augmentation techniques can improve recognition accuracy, an experiment was designed to map the following variables:

- **Independent variable:** Proportion of augmented data added to the training set
- **Dependent variable:** Model testing accuracy

Before training process began, we set hyperparameters (see Table 1). The model designed to run the experiment is simplified to the following steps:

1. Choose appropriate augmented datasets by data synthesis and traditional augmentation technique after similarity comparison
2. Randomly create a training set with 458 real images (70%) and a testing set with 200 images (30%)
3. Add increasing proportions of augmented data (10%, 25%, 50%, 75%) by data synthesis into training set, and record the model testing accuracy
4. Repeat the former steps for augmented data by traditional augmentation technique
5. Compare the recognition accuracy of both augmentation methods over varying proportions of augmented data
6. Use Grad-CAM heatmap to interpret the unexpected results if happens

## Results

*Overall Similarity Comparison*

Figure 2 showed the comparisons of cosine similarity distribution among the Real (113) vs. Real (113), Synthetic (113) vs. Synthetic (113), and Real (160) vs. Synthetic (160) images. The Real vs. Real distribution is left-skewed with a mean of 0.8467. The range is from about 0.6 to 1.0, and the standard deviation (SD) is 0.0529. The Synthetic vs. Synthetic distribution is also left-skewed, with a mean of 0.6712 and a SD of 0.1289. The data distributes between about 0.2 and 1.0. While the range of Real vs. Synthetic distribution is from about 0.2 to less than 0.9, and the shape is slightly right-skewed. The mean is 0.4774, and the SD is 0.0906 (see Table 2).

*Similarity Comparison between Batches*

Figure 3 showed the cosine similarity distributions of different synthetic batches and real images. Each dataset with blood selected 23 images and each dataset without blood selected 24 images. The Real vs. Synthetic Batch 1 distribution is left-skewed, with a mean of 0.6602. The range is from about 0.4 to less than 0.9, and the SD is 0.0845. The Real vs. Synthetic Batch 2 and Real vs. Synthetic Batch 3 are all slightly right-skewed, with means of 0.4819 and 0.4631, respectively. Their ranges are between 0.2 and 0.8, and their SDs are also similar, about 0.07. For Real vs. Synthetic Batch 2 & 3, the distribution is slightly right-skewed, with a mean of 0.4739. The range is from above 0.2 to less than 0.8, and the SD is 0.0865 (see Table 3).

*Image Augmentation Similarity Comparison*

The original images underwent three augmentation processes, respectively - 90 degrees clockwise rotation, horizontal flip, and contrast enhancement. Figure 4 showed the change in the cosine similarity distribution of each augmented dataset and the comparison to the original one.

In Figure 4, all three distributions of the augmented images are left-skewed. Comparing to the distribution of the original images, there exist small differences in their standard deviation. Images by horizontal flip with a mean of 0.8388, while the means of the other two methods are around 0.7 (see Table 4).

*Performance of Model Accuracy*

We used the synthetic batch 2 & 3 and augmented data by horizontal flip as two training set for the image recognition experiment. In Figure 5, the more proportion of data by horizontal flip added, the higher accuracy generated. It showed the classic augmentation technique increased 6% accuracy in the training process after adding 75% augmented data. Surprisingly, the accuracy increased 3% when adding 10% synthetic images in the training set. But the accuracy went down as adding more synthetic images.

In Figure 6, we replaced 100 real images with synthetic images in the testing set, and the original accuracy of the model is 0.59. Recognition accuracy was improved over varying proportions of augmented data by synthesis. The accuracy improved most by adding 50% synthetic images, while the accuracy has shown a downward trend by adding 75%.

To visually validate where our neural network is looking at is the correct area in the image, we used Grad-CAM with 50 images from the testing set. In Figure 6, the first graph indicated the exact region of interest where the blood patterns in the real images for hemophilia patients, while the rest two synthetic images are predicted wrong by the neural network. The second graph is a synthetic image with blood, but the model predicted as "no blood". For the synthetic image without blood in the third graph, it was predicted as "blood".

## Discussion

Overall, data gathered in the experiment demonstrated that two augmentation methods, data synthesis and classic augmentation technique, can improve accuracy to a certain extent. In other words, these two methods can apply to increase the scope of the training set and help to diagnose rare diseases like hemophilia.

*Similarity of Augmented Data by Two Methods*

Figure 2 indicated the similarities between synthetic images and real images are low. The center of the Real vs. Real distribution is much higher than the Real vs. Synthetic one, but the spread is narrower. The reason why real images are more similar and more concentrated is one image repeats about three or five times in the dataset with small changes that we cannot distinguish in the dataset. In addition to the bias of real images, synthetic images also exist bias. The center of the synthetic distribution is less than the real one but is more spread out. This shows that there are huge differences among different similarity levels of batches. Due to the existing bias for both datasets, they cannot match each other well, resulting in the similarity is not high between synthetic and real images.

We originally expected synthetic batch 2 or 3 images would have the best similarity among the three batches. To our shock, the center of batch 1 is the highest in Figure 3. The possible reason why this situation happens is batch 1 images are selected from real images and modified textures manually, while other two batches' images are automatedly generated by artificial intelligence with different change levels. Thus, the synthetic batch 1 images are with the most similarity to the original dataset. In the experiment, we added proportions of synthetic images to confirm our hypothesis. If the sample size of training images is small, it cannot assist in solving the problem of the lack of data available to train the algorithms. Thus, we combined batch 2 and batch 3 as one training set to increase credibility.

Figure 4 demonstrated the cosine similarity distribution of the augmented images by horizontal flip is the most similar, and the center is the closest to original images. This is reasonable, considering that the horizontal flip will not change images too much. Thus, dataset by horizontal flip is another training set in the experiment. Though the chosen image augmentation methods are widely used to artificially increase the number of training images, other affine transformation methods would potentially have closer similarities to the original dataset, such as shearing, scaling, and shifting. Thereby, the result of choosing the most similar dataset is not representative in all image augmentation methods.

*Performance of Model in the Training Process*

Figure 5 displayed model accuracy improved when increasing proportions of augmentation images by horizontal flip added in the training set. Since augmentation images and real images have high similarities, the accuracy would successfully increase. Therefore, augmented data by the traditional augmentation method can be used to increase the training set when the data is limited, which confirmed the $2^{nd}$ hypothesis.

On the other hand, the accuracy went down happened in adding more than 10% synthetic images, because the similarity between synthetic and real images is not good enough, which affects the performance of training the model. For the trained EfficientNet-B4 to be capable of classifying the images, it uses pattern recognition to compare the features it is observing with those it has already seen in order to classify the image into either the "blood" or "no blood" category. This way, the model might ignore some features and areas of images which it thinks it's unimportant or recognize some other areas as the "blood" category. Figure 7 figured out the possible reason why the accuracy is declining with additional synthetic data. The CNN model fails to identify the blood features in synthetic images since the model flounders to determine the exact area of interest and learns something unique to the synthetic data. The synthetic image with blood is predicted as "blood", and the synthetic image without blood is predicted as "no blood". There is a shift away from the domain of interest as increasing the proportions of synthetic data, which causes the loss of accuracy in model training.

Figure 6 showed us that the model accuracy improved in the training process, which means our 1st hypothesis was confirmed. Though there is a downward trend as adding much more augmented images by horizontal flip, the result is reasonable, considering that the size of the training set is enough for training and get great accuracy. If we augment much more images than need, it will affect the model performance negatively. In this research, traditional augmentation data has a better performance than synthetic data in image recognition by increasing the training set. Though it failed to prove the 3rd hypothesis, the different performances will be displayed if we use other datasets.

*Limitation*

Although this research showed two augmentation methods can improve the model accuracy, the limitation of dataset will affect the results. Synthetic and real images are dissimilar to each other which have an opposite impact on model accuracy improvement. Synthetic images with blood and without blood have correlations one to one, and only thing change is the texture inside the region of interest, but real images do not have correlations. Thus, if we have other different datasets, the results of similarity comparison and performance of accuracy will change.

*Future Work*

While this research managed to achieve its objectives, future work could consider using other distance metrics, such as Euclidean distance and Siamese network, for similarity comparison, and compare different methods of measuring image similarities. Additionally, it could repeat with other CNN models for image recognition and see which method, data synthesis or traditional augmentation technique, has a better performance. More importantly, we would want to explore why synthetic data was being recognized by the CNN as different from real data.

**Conclusion**

Similarities between images by cosine similarity and a pre-trained CNN model are assessed and the effect of model accuracy in two augmentation methods, data synthesis and traditional augmentation techniques, are explored. When limited training data is available, our research demonstrated that two augmentation methods can improve accuracy to a certain extent, which confirms 1st and 2nd hypotheses. In addition, the traditional augmentation technique has a better performance than synthesis in the training process, which fails to confirm 3rd hypothesis due to the limitation of datasets. Although future work is still needed to explore the effect of model accuracy when combining synthetic images and augmentation images, introducing augmented data by using data synthesis or traditional augmentation techniques can help to improve the model accuracy, helping in diagnosing rare diseases.

**Acknowledgements**

I would like to express my appreciation to Prof. Pascal Tyrrell, who guided my research. I would also like to thank Mauro Mendez for helping in debugging the codes and allowing me to use his codes for model and Grad-CAM classifications.

# Figures and Tables

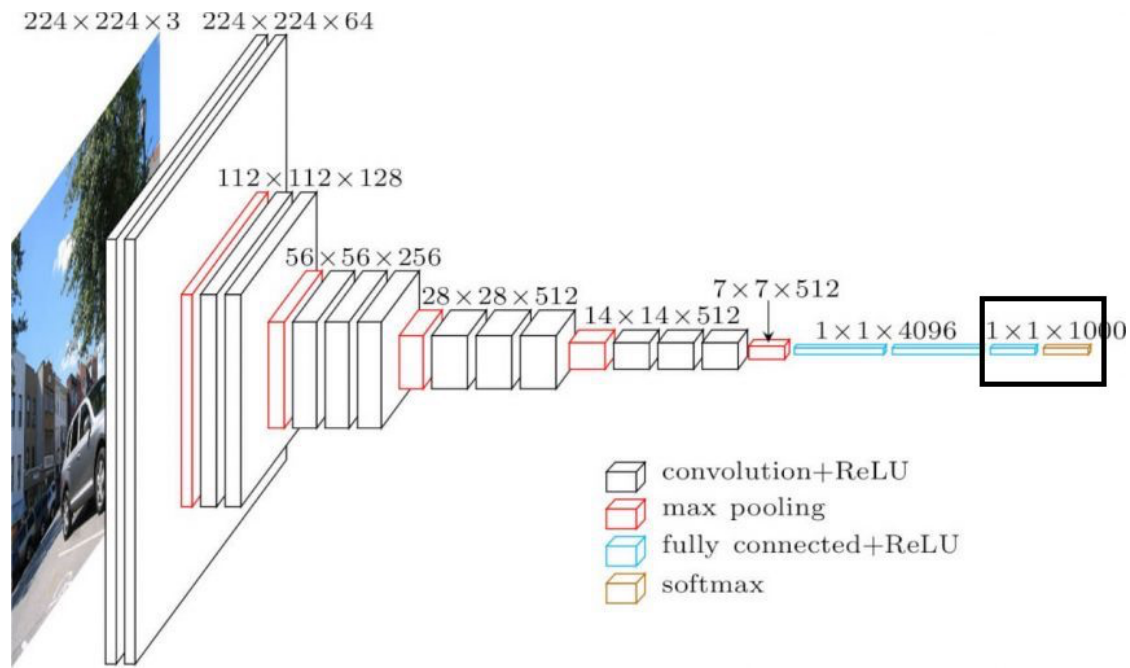

***Figure 1.*** *The process of VGG-16 model. The black box shows these two layers what are not included for feature extraction [12].*

**Table 1:** Hyperparameters for EfficientNet-B4

| Hyperparameter Name | Hyperparameter Value | Description |
| --- | --- | --- |
| **img_size** | 380 | Image input size |
| **batch_size** | 25 | Input batch size for training |
| **data_mean** | [0.1129, 0.1157, 0.1180] | Mean values for each layer (RGB) |
| **data_std** | [0.1546, 0.1575, 0.1595] | Std Dev values for each layer (RGB) |
| **out_features** | 1 | Binary is 1 (with blood or no blood) |
| **optimizer** | ADAM | Optimizer to update model weights |
| **learn_rate** | 0.001 | Learning Rate to use |
| **min_epochs** | 1 | Min number of epochs to train for |
| **epochs** | 50 | Number of epochs to train for |
| **precision** | 16 | PyTorch precision in bits |

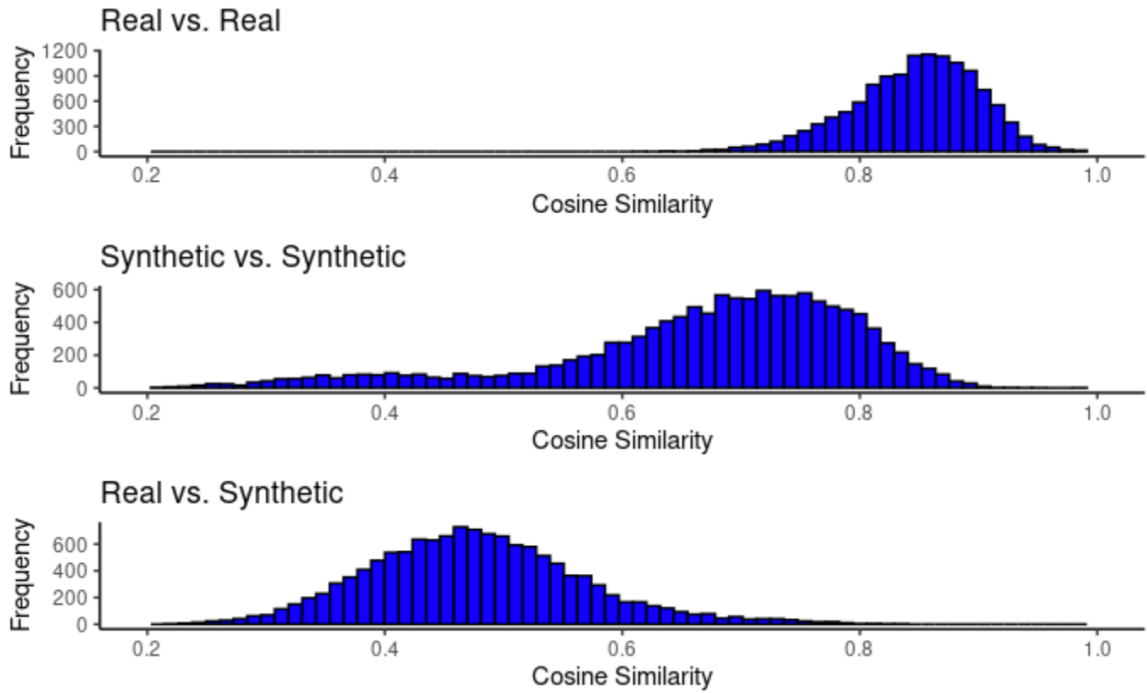

***Figure 2.*** *Comparison of cosine similarities distribution among Real vs. Real, Synthetic vs. Synthetic, and Real vs. Synthetic images. They all compared with blood and without blood, respectively.*

**Table 2:** Summary of data about Figure 2

|  | **Real vs. Real** | **Syn. vs. Syn.** | **Real vs. Syn.** |
|---|---|---|---|
| **Sample Size** | 12656 | 12656 | 12800 |
| **Mean** | 0.8467 | 0.6712 | 0.4737 |
| **SD** | 0.0529 | 0.1289 | 0.0906 |
| **Skewness** | -0.4884 | -1.0920 | 0.4268 |

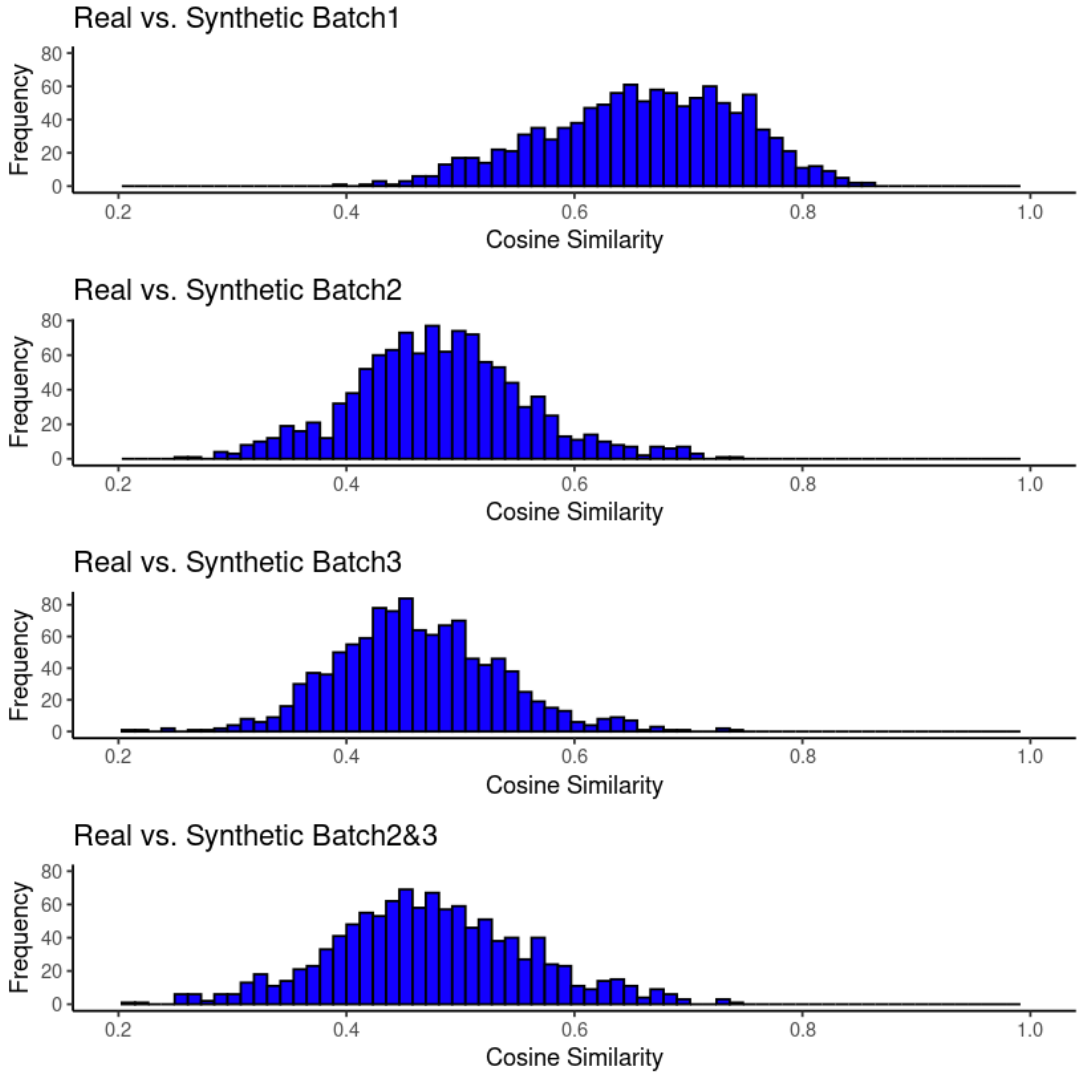

***Figure 3.*** *Comparison of cosine similarities distribution among Real vs. Different Batches of Synthetic images. They all compared with blood and without blood, respectively.*

**Table 3:** Summary of data about Figure 3

|  | Real vs. Syn. Batch 1 | Real vs. Syn. Batch 2 | Real vs. Syn. Batch 3 | Real vs. Syn. Batch 2&3 |
|---|---|---|---|---|
| **Sample Size** | 1105 | 1105 | 1105 | 1105 |
| **Mean** | 0.6602 | 0.4819 | 0.4631 | 0.4739 |
| **SD** | 0.0845 | 0.0771 | 0.0723 | 0.0865 |
| **Skewness** | -0.3111 | 0.2388 | 0.3006 | 0.1252 |

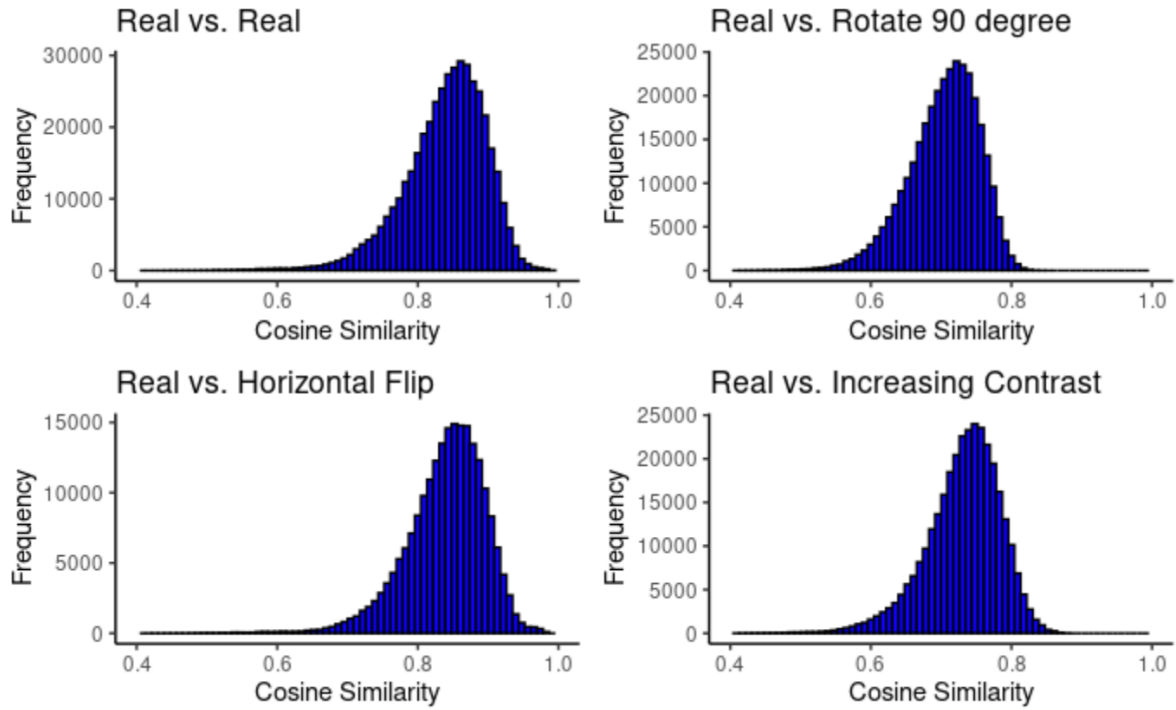

*Figure 4.* Comparison of cosine similarities distribution among three traditional augmentation techniques and original dataset. They all compared with blood and without blood, respectively.

**Table 4:** Summary of data about Figure 4

|  | Original (Reference) | Rotate 90 deg. | Horizontal Flip | Increase Contrast |
|---|---|---|---|---|
| **Mean** | 0.8389 | 0.7034 | 0.8388 | 0.7308 |
| **SD** | 0.0596 | 0.0504 | 0.0575 | 0.0535 |
| **Skewness** | -1.0028 | -0.7418 | -1.0895 | -0.8801 |

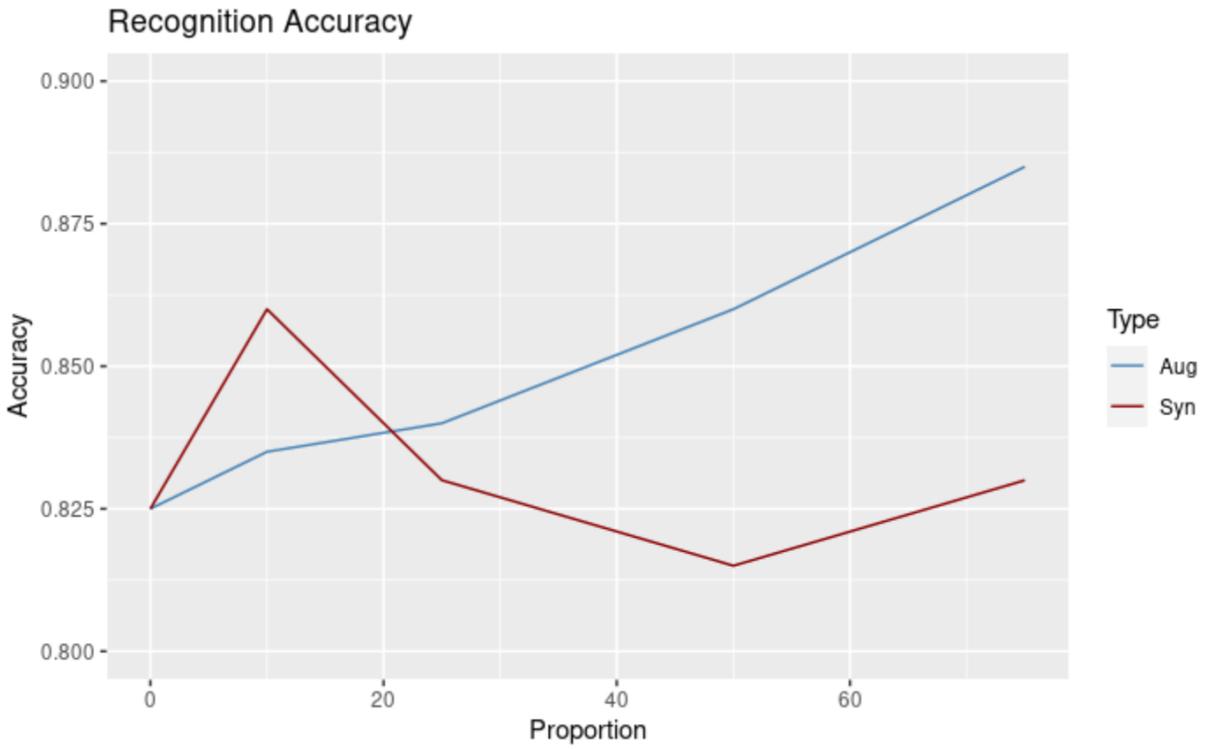

*Figure 5.* Recognition accuracy of both augmentation methods over varying proportions of augmented data (10, 25, 50, 75%). Testing set contains 200 real images.

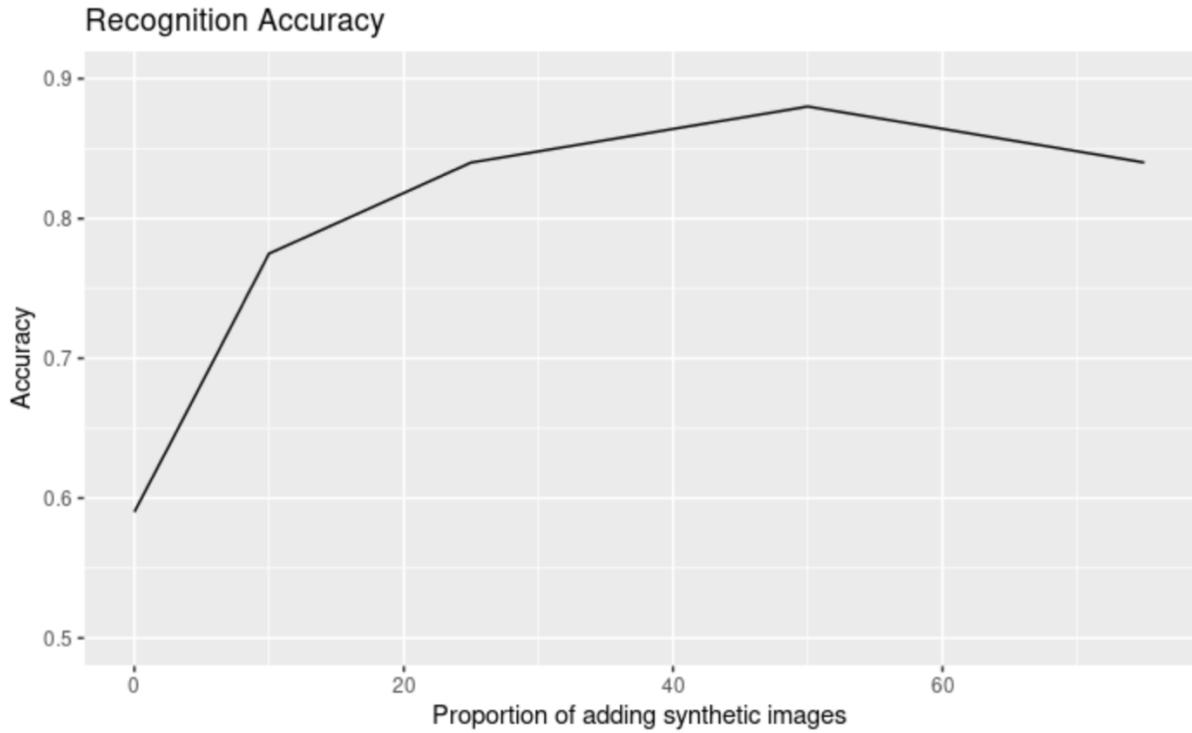

*Figure 6.* Recognition accuracy of augmented data by data synthesis over varying proportions of augmented data (10, 25, 50, 75%). Testing set contains 100 real images and 100 synthetic images.

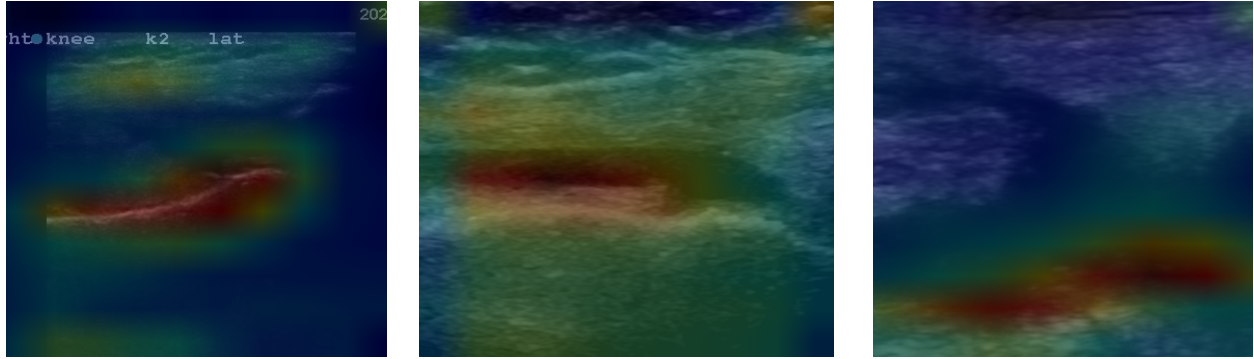

*Figure 7.* The Grad-CAM results for one real image and two synthetic images. The first image is real image with blood and the model also predicted as "blood". The second image is synthetic image with blood, but the model predicted as "no blood", while the last image is also synthetic image without blood, but the model predicted as "blood".